\documentclass{article}

\PassOptionsToPackage{numbers,compress}{natbib}

\usepackage[final]{neurips_2019}

\usepackage[utf8]{inputenc} 
\usepackage[T1]{fontenc}    
\usepackage{hyperref}       
\usepackage{url}            
\usepackage{booktabs}       
\usepackage{amsfonts}       
\usepackage{nicefrac}       
\usepackage{microtype}      
\usepackage{graphicx}
\usepackage{multirow}
\usepackage{mathtools}
\usepackage{nccmath}
\usepackage{amsmath}

\DeclarePairedDelimiter{\nint}\lfloor\rceil

\title{Q8BERT: Quantized 8Bit BERT}

\author{%
  Ofir Zafrir\hspace{\fill}Guy Boudoukh\hspace{\fill}Peter Izsak\hspace{\fill}Moshe Wasserblat\\
  Intel AI Lab\\
  \texttt{\{ofir.zafrir, guy.boudoukh, peter.izsak, moshe.wasserblat\}@intel.com} \\
}
\begin{document}
\maketitle

\begin{abstract}
    Recently, pre-trained Transformer \cite{vaswani2017attention} based language models such as BERT \cite{Devlin2018BERTPO} and GPT \cite{radford2018improving}, have shown great improvement in many Natural Language Processing (NLP) tasks.
    However, these models contain a large amount of parameters.
    The emergence of even larger and more accurate models such as GPT2 \cite{radford2019language} and Megatron\footnote{\url{https://github.com/NVIDIA/Megatron-LM}}, suggest a trend of large pre-trained Transformer models. 
    However, using these large models in production environments is a complex task requiring a large amount of compute, memory and power resources. 
    In this work we show how to perform quantization-aware training during the fine-tuning phase of BERT in order to compress BERT by $4\times$ with minimal accuracy loss. 
    Furthermore, the produced quantized model can accelerate inference speed if it is optimized for 8bit Integer supporting hardware.
\end{abstract}

\section{Introduction} \label{sec:intro}
Pre-trained transformer language models (GPT \cite{radford2018improving}, XLNet \cite{yang2019xlnet}, XLM \cite{lample2019cross}, BERT \cite{Devlin2018BERTPO}) have demonstrated State-of-the-Art (SOTA) results for a variety of NLP tasks such as sentence classification, sequence tagging and question answering, by extracting contextual word representations or by fine-tuning the whole model on a target task.
The models are pre-trained on extremely large corpora and result in a large number of parameters.
For example, \citet{Devlin2018BERTPO} introduced two pre-trained models: BERT-Base, which has 110M parameters in 32bit Floating Point (FP32) representation, and BERT-Large, which has 334M parameters in FP32 representation.
Both BERT models have a high memory footprint and require heavy compute and wide bandwidth during inference.
In addition, real time NLP applications that integrate BERT have to meet low latency requirements to achieve a high quality customer experience, therefore, the computational characteristics of BERT pose a challenge to deployment in production environments.
These models will have a major impact on the way business organizations consume computing resources, since computing resources will have to handle loading of large models and heavy feed-forward calculations, shifting workload focus from lower level training to more application-specific fine-tuning and inference. 
Therefore, it is crucial to develop energy-efficient and minimum-cost methods to run these models in production \cite{moshe2019nlpfuture}.

Model compression algorithms are used to reduce compute and memory resources required for running inference. 
For example, \citet{han2015deep} used a pipeline of pruning, quantization and Huffman encoding in order to achieve a compression ratio of $49\times$ of VGG-16 \cite{simonyan2014very}. 
As a result, the compressed VGG-16 can be fitted into an on-chip SRAM cache which allows faster access times with less power in comparison to off-chip DRAM memory. 
In another example, \citet{jacob2018quantization} introduced a method of training linear quantized Convolutional Neural Networks (CNN) that uses Integer Arithmetic instead of Floating Point Arithmetic which can be up to $4\times$ faster using only $25\%$ of the memory footprint \cite{37631}.

In this work, we present a method for achieving best-in-class compression-accuracy ratio for BERT.
To do this, we apply quantization-aware training during the fine-tuning process of BERT.
We quantize all GEMM (General Matrix Multiply) operations in BERT Fully Connected (FC) and Embedding layers. We simulate 8bit quantized inference while maintaining $99\%$ accuracy in comparison to the FP32 version of BERT for eight different NLP tasks.
Moreover, since we quantize all the FC and Embedding layers' weights - which comprise over $99\%$ of the model's weights - to 8bit, we achieve a memory footprint $4\times$ smaller than the original BERT. 
In addition, it is possible to use our method to implement efficient inference with hardware that supports 8bit arithmetic and an optimized library for 8bit GEMM.
We have released our work as part of our open source model library \texttt{NLP Architect}\footnote{\url{https://github.com/NervanaSystems/nlp-architect}}.

The method presented in this paper is not exclusive to BERT model and can be integrated into other large pre-trained Transformer based models.

\section{Method} \label{sec:method}
In this section, we describe the quantization scheme, linear quantization, and quantization-aware training method we used.
We chose to use this quantization scheme because, in addition to reducing the model size by approximately $4\times$, it is also possible to accelerate inference time by using Integer arithmetic to calculate GEMM using specialized hardware for Integer and Fixed Point calculations. 
For example, \citet{bhandare2019efficient} stated that using Intel\textsuperscript{\textregistered} Xeon\textsuperscript{\textregistered} Cascade Lake's Vectorized Neural Network Instructions (VNNI) to perform Int8 matrix multiplication provides a speed-up of $3.7\times$ over FP32 matrix multiplication.
Moreover, by using symmetric linear quantization we simplify the quantization process and zero out terms related to the offset part of the quantized values.
Our method is based on the method proposed by \citet{jacob2018quantization}.

\subsection{Quantization Scheme} \label{sec:scheme}
We use symmetric linear quantization as our quantization scheme for quantizing both weights and activations to 8bit Integers (Int8):
\begin{equation}
  \begin{aligned}
    \text{Quantize}(x|S^{x}, M):=\text{Clamp}\left (\nint{x\times S^{x}}, -M, M \right ),\\
    \text{Clamp}\left (x, a, b\right )=\min{\left(\max{\left(x,a\right)},b\right)}
    \end{aligned}
\end{equation}
where $S^{x}$ is the quantization scaling-factor for input $x$ and $M$ is the highest quantized value when quantizing to $b$ number of bits:
\begin{equation}
    M=2^{b-1}-1
\end{equation}
E.g. when quantizing to 8 bits, $M=127$. 
The scaling-factor can be determined either dynamically during inference, or calculated using statistics collected during training, or calculated using statistics collected, post-training, during inference on a calibration set. 
In our work the weights' scaling-factor is calculated according to:
\begin{equation}
\label{eq:wscale}
    S^{W}=\frac{M}{\max\left(\left|W\right|\right)}
\end{equation}
and the activations' scaling-factor is calculated based on values seen during training using an Exponential Moving Average (EMA):
\begin{equation}
\label{eq:ascale}
    S^{x}=\frac{M}{\text{EMA}\left(\max\left(\left|x\right|\right)\right)}
\end{equation}

\subsection{Quantized-Aware Training} \label{sec:quant_aware}
Quantization-aware training is a method of training Neural Networks (NN) to be quantized at the inference stage, as opposed to post-training quantization where the training is executed without any adaptation to the quantization process.
In our work, we use fake quantization to introduce the quantization error to the model during the training phase in order for the model to learn to bridge the quantization error gap.
Fake quantization is an operation that simulates the rounding effect in Floating Point values as presented by \citet{jacob2018quantization}.
Since the rounding operation is not derivable, we use the Straight-Through Estimator (STE) \cite{bengio2013estimating} to estimate the gradient of fake quantization:
\begin{equation}
    \frac{\partial x^{q}}{\partial x}=\overrightarrow{\textbf{1}}
\end{equation}
where $x^{q}$ denotes the result of fake quantizing $x$.
Using the combination of fake quantization and STE we are able to perform quantized inference during training while back-propagating at full precision which allows the FP32 weights to overcome the quantization error.

\section{Implementation}
Our goal is to quantize all the Embedding and FC layers in BERT to Int8 using the method described in Section~\ref{sec:method}. 
For this purpose we implemented quantized versions of Embedding and FC layers. 
During training, the Embedding layer returns fake quantized embedding vectors, and the quantized FC performs GEMM between the fake quantized input and the fake quantized weight, and then accumulates the products to the bias which is untouched since the bias will be later quantized to Int32.
During inference, the quantized Embedding layer returns Int8 embedding vectors, and the quantized FC performs GEMM between Int8 inputs accumulated to the Int32 bias which is quantized using the weights' and activations' scaling-factors as described in \cite{jacob2018quantization}. 
Although the bias vectors are quantized to Int32 values, they only make up for a fraction of the amount of parameters in the model.

Our implementation of Quantized BERT is based on the BERT implementation provided by the \texttt{PyTorch-Transformers}\footnote{\url{https://github.com/huggingface/pytorch-transformers}} library. 
To implement quantized BERT we replaced all the Embedding and FC layers in BERT to the quantized Embedding and FC layers we had implemented. 
Operations that require higher precision, such as Softmax, Layer Normalization and GELU, are kept in FP32.

\begin{table}[t]
\caption{GLUE tasks and SQuAD results. Each score is evaluated on the publicly available development set for the task using the metric specified for each task. For each task we present the score of a baseline (FP32) model, of a Quantization-Aware Training (QAT) model quantized to 8bit, and of a Dynamically Quantized (DQ) to 8bit model. Large means those tasks were trained with BERT-Large architecture.}
\label{tab:results}
\centering
\begin{tabular}{llccc}
\hline
Dataset     & Metric          & \begin{tabular}[c]{@{}l@{}}BERT baseline \\ accuracy (STD)\end{tabular} & \begin{tabular}[c]{@{}l@{}}QAT BERT \\ 8bit (STD)\end{tabular}    & \begin{tabular}[c]{@{}l@{}}DQ BERT \\ 8bit (STD)\end{tabular} \\ \hline
CoLA        & Matthew's corr. & \textbf{58.48} (1.54)                                                   & \textbf{58.48} (1.32)                                             & 56.74 (0.61)        \\
MRPC        & F1              & \textbf{90} (0.23)                                                      & 89.56 (0.18)                                                      & 87.88 (2.03)        \\
MRPC-Large  & F1              & 90.86 (0.55)                                                            & \textbf{90.9} (0.29)                                              & 88.18 (2.19)        \\
QNLI        & Accuracy        & 90.3 (0.44)                                                             & \textbf{90.62} (0.29)                                             & 89.34 (0.61)        \\
QNLI-Large  & Accuracy        & 91.66 (0.15)                                                            & \textbf{91.74} (0.36)                                             & 88.38 (2.22)        \\
QQP         & F1              & 87.84 (0.19)                                                            & \textbf{87.96} (0.35)                                             & 84.98 (0.97)        \\
RTE         & Accuracy        & \textbf{69.7} (1.5)                                                     & 68.78 (3.52)                                                      & 63.32 (4.58)        \\
SST-2       & Accuracy        & \textbf{92.36} (0.59)                                                   & 92.24 (0.27)                                                      & 91.04 (0.43)        \\
STS-B       & Pearson corr.   & \textbf{89.62} (0.31)                                                   & 89.04 (0.17)                                                      & 87.66 (0.41)        \\
STS-B-Large & Pearson corr.   & \textbf{90.34} (0.21)                                                   & 90.12 (0.13)                                                      & 83.04 (5.71)        \\
SQuADv1.1   & F1              & \textbf{88.46} (0.15)                                                   & 87.74 (0.15)                                                      & 80.02 (2.38)        \\ \hline
\end{tabular}%
\end{table}

\section{Evaluation}
To test our approach we evaluated our model on the GLUE (General Language Understanding Evaluation) benchmark \cite{wang2018glue}, which is a collection of resources for training, evaluating, and analyzing natural language understanding systems in a wide array of NLP tasks. 
The ultimate goal of GLUE is to drive research in the development of general and robust natural language understanding systems. 
In addition, we evaluated our model on the question and answering task SQuADv1.1 \cite{rajpurkar2016squad}. 
The Stanford Question Answering Dataset (SQuAD) is a reading comprehension dataset, consisting of questions posed by crowd workers on a set of Wikipedia articles, where the answer to every question is a segment of text, or span, from the corresponding reading passage.

We summarized our results for quantized BERT in the QAT column in Table~\ref{tab:results}. 
We ran each experiment five times and reported the average result and standard deviation. 
In addition, we calculated the relative error induced by the quantization process and summarized the results in Table~\ref{tab:error}.
In all experiments we used BERT-Base as the base model unless indicated otherwise. 
In all experiments we fine-tuned the pre-trained models offered by Tensorflow-Hub\footnote{\url{https://www.tensorflow.org/hub}}. 
In our internal testing, we found that the relative error induced by quantization is less than $1\%$ (excluding RTE task) while the space capacity of the model is reduced by 4x.

\subsection{Effect of Quantization-Aware Training} \label{sec:qat-effect}
In order to measure the necessity of quantization-aware training we compared our results to post-training quantized models. 
We quantized our baseline models using Dynamic Quantization (DQ).
The weights and activations are quantized as described in Section~\ref{sec:scheme} with a small difference in the way we calculate the quantization scaling-factor of the activations. 
Instead of using Equation~\ref{eq:ascale} we compute the scale the same way we compute the weights' scaling-factor using Equation~\ref{eq:wscale}. 
This calculation is done during inference for each incoming activation tensor. 
The results for this quantization method are summarized in the DQ column in Table~\ref{tab:results} and the relative error induced by quantization is also summarized in Table~\ref{tab:error}. 
We observe that the DQ method produces significantly worse results over all tasks. 

\begin{table}[t]
\caption{Reduction in accuracy induced by quantization relative to baseline model. DQ is the Dynamically Quantized model and QAT is the Quantization-aware Trained quantized model. Large means those tasks were trained with BERT-Large architecture.}
\label{tab:error}
\resizebox{\textwidth}{!}{%
\begin{tabular}{cccccccccccc}
\hline
Method & CoLA   & MRPC   & MRPC-Large & QNLI    & QNLI-Large & QQP     & RTE    & SST-2  & STS-B  & STS-B-Large & SQuADv1.1 \\ \hline
DQ & 2.98\% & 2.36\% & 2.95\%     & 1.69\%  & 3.58\%     & 1.85\%  & 9.13\% & 1.43\% & 2.19\% & 8.08\%      & 9.54\%    \\
QAT & 0.00\% & 0.49\% & -0.04\%    & -0.35\% & -0.09\%    & -0.14\% & 1.32\% & 0.13\% & 0.65\% & 0.24\%      & 0.81\%    \\ \hline
\end{tabular}%
}
\end{table}

\section{Related Work}
Compressing Transformer-based models for efficient inference is an active field of research. 
\citet{junczys2018marian} applied knowledge distillation and 8bit post-training quantization to speed up Transformer models for neural machine translation (Transformer-LT) \cite{vaswani2017attention}, however, the quantized model suffered a loss of 1 BLEU score in comparison to the baseline model. 
\citet{bhandare2019efficient} also applied 8bit post-training quantization to Transformer-LT models and demonstrated how to utilize Intel\textsuperscript{\textregistered} specialized 8bit hardware to accelerate the inference process.
Habana Labs\footnote{\url{https://habana.ai/habana-labs-goya-delivers-inferencing-on-bert/}} published Quantized BERT performance measurements on its in-house accelerator for NN inference, however, Habana quantized BERT to 16bit Integer which offers a much wider quantization range and only $2\times$ compression. 
NVIDIA\footnote{\url{https://devblogs.nvidia.com/nlu-with-tensorrt-bert/}} also measured BERT performance on its in-house accelerator using 16bit Floating Point arithmetic. 
Furthermore, NVIDIA implemented a number of optimized kernels for BERT's operations in order to save memory bandwidth during inference. 
\citet{sam2019compressbert} fine-tuned BERT on a custom dataset and performed 8bit Integer post-training quantization.

\section{Conclusions and Future Work}
We have shown a method for quantizing BERT GEMM operations to 8bit for a variety of NLP tasks with minimum loss in accuracy, and hope that the software developers community can use our quantization method to compress BERT and implement efficient BERT inference with 8bit GEMM operations. 
Efficient inference will enable low-latency NLP applications on a variety of hardware platforms from edge devices to data centers. 
In the future we intend to apply other model compression methods in order to compress BERT. 
Decreasing BERT’s memory footprint will accelerate BERT inference time and reduce power consumption, both of which are critical for deploying
BERT in production environments having low memory and power resources.
Furthermore, we intend to integrate other compression methods with our quantized BERT model.



\bibliographystyle{abbrvnat}
\bibliography{references}

\end{document}